\documentclass[10pt,twocolumn,letterpaper]{article}

\usepackage{wacv}
\usepackage{times}
\usepackage{epsfig}
\usepackage{graphicx}
\usepackage{amsmath}
\usepackage{amssymb}
\usepackage{bm}

\usepackage{microtype}
\usepackage{graphicx}
\usepackage{subfigure}
\usepackage{booktabs}

\usepackage{amsfonts}
\usepackage{import}
\usepackage{float}
\usepackage{algorithm}
\usepackage{algorithmic}
\usepackage{xspace}
\usepackage{authblk}

\usepackage[symbol]{footmisc}

\usepackage[dvipsnames]{xcolor}

\def\wacvPaperID{280} 

\wacvfinalcopy

\ifwacvfinal
\fi

\newcommand{\fidelity}{\textit{Fidelity}\xspace}
\newcommand{\stability}{\textit{Stability}\xspace}
\newcommand{\comprehensibility}{\textit{Comprehensibility}\xspace}
\newcommand{\consistency}{\textit{Consistency}\xspace}
\newcommand{\representativity}{\textit{Generalizability}\xspace}
\newcommand{\mR}{MeGe\xspace}
\newcommand{\mC}{ReCo\xspace}

\newcommand{\Sp}{\boldsymbol{\mathcal{S}^{ \ne }}}
\newcommand{\Sm}{\boldsymbol{\mathcal{S}^{ = }}}
\newcommand{\Sa}{\boldsymbol{\mathcal{S}}}

\newcommand{\D}{\mathcal{D}}
\newcommand{\X}{\mathcal{X}}
\newcommand{\Y}{\mathcal{Y}}
\newcommand{\vz}{\bm{z}}
\newcommand{\vx}{\bm{x}}
\newcommand{\vy}{\bm{y}}
\newcommand{\V}{\mathcal{V}}
\newcommand{\A}{\mathcal{A}}

\newcommand{\f}{\bm{f}}
\newcommand{\G}{\bm{\Phi}}
\newcommand{\g}{\bm{\phi}}
\newcommand{\dg}{\delta_{\vx}^{(i,j)}}

\definecolor{blue}{RGB}{71, 106, 227}
\definecolor{red}{RGB}{234, 71, 47}
\definecolor{yellow}{RGB}{248, 145, 0}
\definecolor{indigo}{RGB}{63, 81, 181}

\newtheorem{definition}{Definition}

\ifwacvfinal
\usepackage[breaklinks=true,bookmarks=false,colorlinks,citecolor=indigo]{hyperref}
\else
\usepackage[pagebackref=true,breaklinks=true,colorlinks,bookmarks=false,citecolor=indigo]{hyperref}
\fi

\title{How Good is your Explanation? Algorithmic Stability Measures to Assess the Quality of Explanations for Deep Neural Networks}

\begin{document}

\author[1,2,3]{Thomas Fel}
\author[3]{David Vigouroux}
\author[1]{Rémi Cadène}
\author[1,2]{Thomas Serre}
\affil[1]{Carney Institute for Brain Science, Brown University}
\affil[2]{Artificial and Natural Intelligence Toulouse Institute, Université de Toulouse, France}
\affil[3]{IRT Saint-Exupery}
\maketitle
\thispagestyle{empty}

\footnotetext{Email correspondance to: \tt\small{thomas\_fel@brown.edu}}

\vspace{-2mm}\begin{abstract}
A plethora of methods have been proposed to explain how deep neural networks reach their decisions but comparatively, little effort has been made to ensure that the explanations produced by these methods are objectively relevant. While several desirable properties for trustworthy explanations have been formulated, objective measures have been harder to derive. Here, we propose two new measures to evaluate explanations borrowed from the field of algorithmic stability: mean generalizability \mR and relative consistency \mC. We conduct extensive experiments on different network architectures, common explainability methods, and several image datasets to demonstrate the benefits of the proposed measures.
In comparison to ours, popular fidelity measures are not sufficient to guarantee trustworthy explanations.
Finally, we found that 1-Lipschitz networks produce explanations with higher \mR and \mC than common neural networks while reaching similar accuracy. This suggests that 1-Lipschitz networks are a relevant direction towards predictors that are more explainable and trustworthy.
\end{abstract}
\vspace{-3mm}

\section{Introduction}
\label{introduction}
Machine learning approaches such as deep neural networks have become essential in multiple domains such as image classification, language processing and speech recognition. These approaches have achieved excellent classification accuracy -- approaching human performance in specific domains~\cite{lecun2015deep, Serre2019DeepLT}.
However, one significant drawback associated with these deep networks is that it is difficult to interpret their decisions~\cite{lipton2016mythos}. This problem constitutes a serious obstacle for the wide adoption of these systems for safety-critical applications.

\begin{figure}[t!]
  \includegraphics[width=0.47\textwidth]{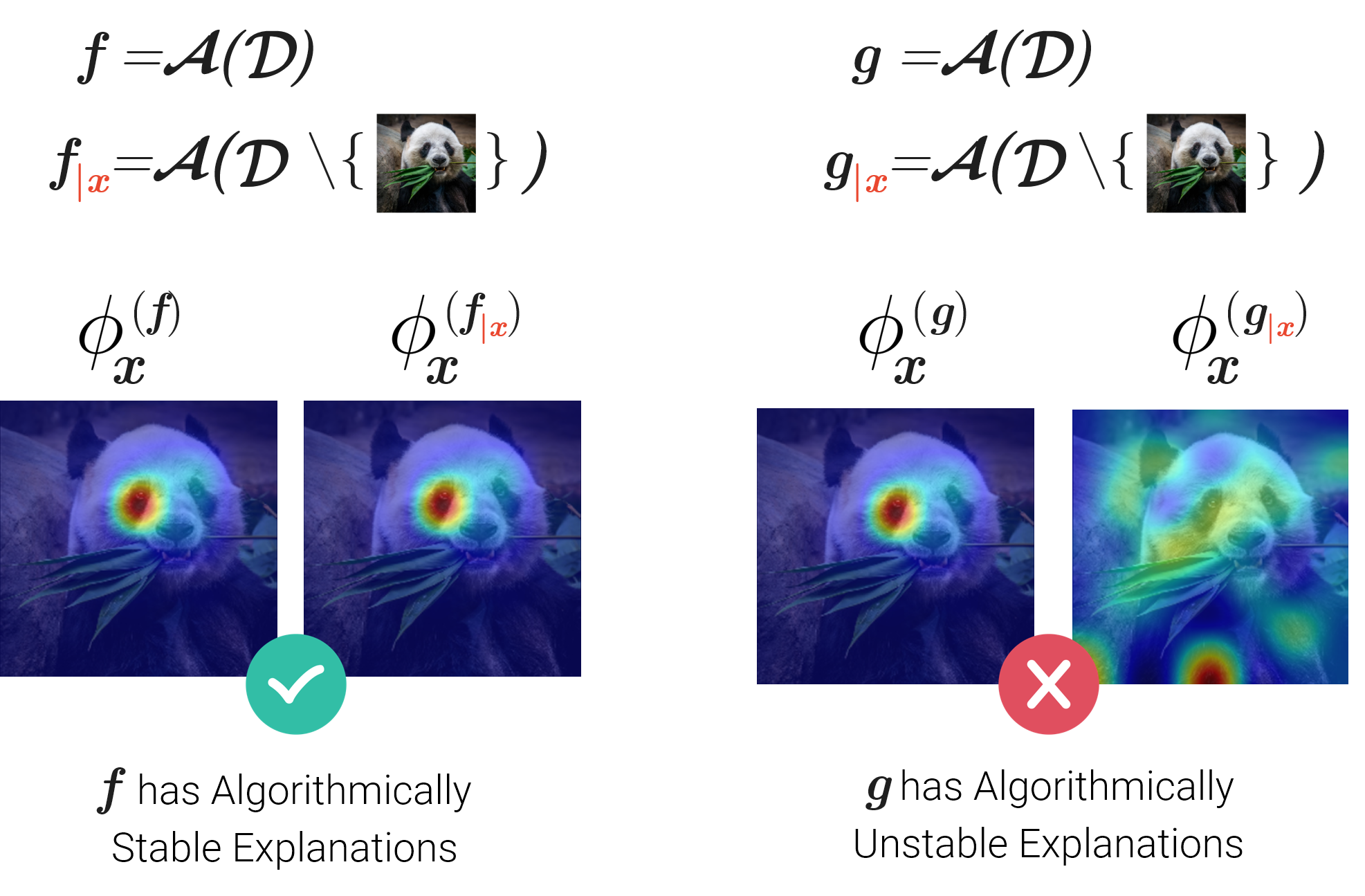}
  \caption{
  The explanations of a predictor $\f$ can be trusted when they are algorithmically stable. It means that, on average, for any image $\vx$ removed for the training set $\D$, another predictor $\f_{\textcolor{red}{|\vx}}$ trained with the same algorithm $\A$ produces a similar explanation for the same input image. This stability ensures that the explanations of correctly classified images of the same class will point out similar evidence of the class. For instance, the explanations for the class \textit{panda} will all point out the black area around their eyes.
  }
  \label{fig:big_picture}
  \vspace{-5mm}
\end{figure}

Recently, several explainability methods have been proposed to help understand how predictors make a particular decision~\cite{sundararajan2017axiomatic, smilkov2017smoothgrad, petsiuk2018rise, Selvaraju_2019}.
These methods are also meant to improve predictors and more importantly to build trust with people affected by their predictions. 
Unfortunately, these methods have strong limitations. In particular, they are subject to confirmation bias: while some methods appear to offer useful explanations to a human experimenter, they turn out not to reflect the actual behaviour of the predictor~\cite{adebayo2018sanity, ghorbani2017interpretation}.
Instead of providing confidence in a system's decisions, these explanations can themselves be potentially erroneous and cannot be trusted.

In this work, we focus on evaluation methods that objectively assess the quality and trustworthiness of explanations.
Numerous methods have been proposed~\cite{survey2019metrics, tintarev2007survey, miller2017explanation, robnik2018perturbation, gilpin2018explaining, alvarezmelis2018robust}. They are often derived from axioms and properties that good explanations should possess.
The most studied property is \fidelity. It is associated with a metric that evaluates an explanation method on its ability to reflect the internal decision process of a given neural network predictor. An important limitation of \fidelity is that it does not take into account the prediction capability of the predictor. For instance, an explainability method could have a high \fidelity given a random predictor. Thus, \fidelity is only a first step towards predictors that are explainable and can be trusted.

Instead, we propose to additionally take the prediction capability of the predictor into account through the \representativity and \consistency of its explanations.
A shown in Figure~\ref{fig:big_picture}, we borrow these two concepts from the fields of generalization and algorithmic stability~\cite{bousquet2002stability}, and adapt them to the field of explainability.
Intuitively, when some images of a certain class are well classified by a given predictor, trustworthy explanations should point out to the same evidence accross all images of the same category. For instance, the black region near the eyes of a panda. That is captured by the notion of \representativity.
Conversely, when images of the same category are misclassified, trustworthy explanations should ideally point out to image evidence which differs from that used for correctly classified images. That is captured well by the notion of  \consistency.
In practice, we estimate with a $k$-\textit{fold} cross-training approach the metrics associated with different predictors, i.e., their average generalizability (\mR) and their relative consistency (\mC).



We provide an extensive experimental validation of our approach using different neural networks, common explainability methods and several image datasets. We compare the average generalizability (\mR) and relative consistency (\mC) measures against the leading \fidelity~ measure~\cite{aggregating2020, yeh2019infidelity}.
Importantly, we experimentally show that the \fidelity is not enough to ensure that a given pretrained model produces trustworthy explanations. Finally, we show that 1-Lipschitz networks obtain a higher average generalizability (\mR) and relative consistency (\mC) compared to common neural networks while reaching similar accuracies. This finding offers an interesting research track towards more explainable and trustworthy predictors. To summarize, {\bf our contributions} include:
\vspace{-1.5mm}\begin{itemize}
    \setlength\itemsep{-0.2em}
    \item \representativity~(\mR)~and~\consistency~(\mC) practical measures to assess the quality and trustworthiness of explanations,
    \item Extensive experimental validation of our approach on several neural networks, explainability methods, and image datasets (including ImageNet). 
    \item Empirical demonstration that 1-Lipschitz networks deliver explanations with higher \mR and \mC.
\end{itemize}

\vspace{-4mm}\section{Related work}
\label{related}

In this work, we focus on evaluating explanations provided by explainability methods, which give insight into how a given neural network architecture reaches a particular decision \cite{doshivelez2017rigorous}. These explainability methods produce an influence score for each input dimension. In the case of  image classification, these methods will produce heatmaps indicating the diagnosticity of individual image regions. Most of these explainability methods rely on backpropagating the gradient with respect to a given input image \cite{Zeiler2011, simonyan2013deep, bach2015pixel, Fong_2017, shrikumar2017learning, sundararajan2017axiomatic, smilkov2017smoothgrad, Selvaraju_2019, hartley2021swag} or with respect to a perturbation of the input \cite{zeiler2013visualizing, zhou2014object, ribeiro2016i, li2016understanding, zintgraf2017visualizing, ribeiro2018anchors, fel2021sobol}.

Despite a wide range of explainability methods, assessing the quality and trustworthiness of these explanations is still an open problem. It is in part due to the difficulty of obtaining objective ground truths \cite{samek2015evaluating, linsley2018learning}. Several criteria have been proposed to evaluate the quality of explanations \cite{tintarev2007survey, miller2017explanation, robnik2018perturbation, gilpin2018explaining, alvarezmelis2018robust, survey2019metrics, ferrettini2021coalitional}. According to  \cite{survey2019metrics}, the five major properties include: \fidelity, \stability, \comprehensibility, \representativity~and \consistency. Yet, properties such as \representativity~and \consistency do not come with a practical definition.

In order to measure these different properties, there are two main approaches currently used. The first subjective approach consists in putting the human at the heart of the process, either by explicitly asking for human feedback \cite{Selvaraju_2019, ribeiro2016i, lundberg2017unified}, or by indirectly measuring the performance of the human/classifier duo \cite{lage2019evaluation, NIPS2009_3700, narayanan2018humans, schmidt2019quantifying}. Nevertheless, human intervention sometimes brings undesirable effects, including a possible confirmation bias  \cite{adebayo2018sanity}.

A second type of approaches has also started to emerge specifically for computer vision applications. The main idea is to build objective proxy tasks that a good explanation must be able to solve.
These measures aim to evaluate explanations based on two properties: \fidelity~and \stability. The first method to measure \fidelity~was first proposed in \cite{samek2015evaluating} based on estimating the drop in prediction score resulting from deleting pixels deemed important by an explanation method. To ensure that the drop in score does not come from a change in distribution, ROAR\cite{hooker2018benchmark} was proposed which re-train a classifier model between each deletion step. This boils down to measuring the correlation between the attributions for each pixel and the difference in the prediction score when they are modified and has been clearly formalized \cite{yeh2019infidelity, aggregating2020, rieger2020irof}.
Nevertheless, it should be noted that the different fidelity metrics proposed requires defining a baseline state which might favor explainability methods that internally relies on the same baseline \cite{sturmfels2020visualizing}.

Those \fidelity~metrics are a first step toward trustworthy explanations: by making sure that we have faithful explanations, we can then look at other criteria to quantitatively measure these explanations. 
However, the different \fidelity~metrics suffer from limitations demonstrated by \cite{tomsett2019sanity} which proposes several sanity checks revealing their high variance, their sensitivity to the baseline and their low consistency.

\vspace{-2mm}\section{Methods}
\label{framework}
Below, we briefly provide some motivation for the proposed  \mR~and \mC~ measures before describing a training procedure applicable to a large family of machine learning models in order to estimate these two values. 
One basic assumption for the proposed approach is borrowed from algorithmic stability and generalization: to be reliable, an explanation obtained for a  specific predictor for a given image should be stable when the training dataset used to train the predictor is perturbed slightly as when, for instance, this particular image is added or removed from the dataset. 

\subsection{Notations}

We consider a standard supervised learning setting where a datapoint is denoted $\vz = (\vx, \vy)$ s.t. $\vx \in \X$ is an observation (e.g., $\X = \mathbb{R}^d$) and $\vy \in \Y$ is a class label (e.g., $\Y = \mathbb{R}^p$). 
The data set is denoted as $\D = \{ \vz_1, ..., \vz_m \}$,  we designate $\V = \{ \V_1, ..., \V_k \}$ the set of $k$ disjoints subsets (\textit{folds}) of size $m/k$ at random where each $\V_i \subset \D$. Throughout this work, we will assume $k$ divides $m$ for convenience.
Let $\A$ be a deterministic learning algorithm that maps any number of data points onto a function $\f$ from $\X$ to $\Y$.
In particular, we consider the \textit{fold} $\V_i$ and the associated predictor $\f_i = \A(\V \setminus \V_i)$.

An explanation method is a functional, denoted $\G$, which, given a predictor $\f_i$ and a datapoint $\vx$, assigns an importance score for each input dimension $\g_{\vx}^{(i)} = \G(\f_i, \vx)$.
We define a distance $d(\cdot,\cdot)$ over the explanations.
Finally, the following Boolean connectives are used: $\neg$ denotes a negation, $\land$ denotes a conjunction, and $\oplus$ denotes an exclusive or (XOR).

\subsection{Motivation}
    
We first consider \representativity: we provide a definition, discuss the inherent difficulties associated with its measurement, and describe a method for estimating it.
We then motivate the need for assessing the \consistency~of an explanation and propose a measure.

\begin{definition}{\representativity}\\
A measure of how generalizable an explanation is, and the extent to which it truly reflects the underlying process by which the predictor makes a decision.
\label{def:representativity}
\end{definition}

Intuitively, a representative explanation would be an explanation that holds for a large number of samples.
To assess the number of samples that can be covered by a given explanation, it might be tempting to compute a distance between the explanations associated with those samples. However,  because of the large variations in the appearance of objects that arise because of translation, scale, and 3D rotation in natural images, two explanations can be similar (i.e., close in pixel space) without necessarily reflecting a similar visual strategy used by the predictor (for instance, decisions could be driven by the same pixel locations -- yet driven by different visual features). Conversely, two spatially distant explanations could be based on the same features that appear at different locations because of translation. 

Our proposed solution to this problem is to only use distance measured between explanations for the same sample. This constraint leads us to consider the notion of algorithmic stability as a proxy for generalization: intuitively, given a predictor and a training data set, a good explanation for a decision made for a given data point should be robust to the addition or removal of that data point from the training set. One benefit of such a characteristic is that it can be evaluated based solely on a distance between explanations from the same sample.

In what follows, we will propose a relaxed version of the algorithmic stability -- computationally more manageable -- applied to the explanations using several predictors trained on different \textit{folds}.
It is important to note that the term algorithmic stability \cite{bousquet2002stability} is not related to the \stability~of an explanation as defined in~\cite{aggregating2020}.

Following this consideration, we will be looking at how well a predictor's explanations generalize from seen to unseen data points:

\vspace{-1.5mm}\begin{equation}
    \label{eq:quantity_of_interest}
    \dg = d(\g_{\vx}^i, \g_{\vx}^j) \; s.t. \; \vx \in \V_i, i \neq j.
    \vspace{-1mm}
\end{equation}

\vspace{0.3mm}\noindent By making sure that $\vx$ belongs to the \textit{fold} $\V_i$, we measure the distance between two explanations, one of which comes from a predictor that was not fitted to the sample $\vx$. 
By computing these distances, we hope to characterize the \representativity~of the explanations.

\begin{definition}{\consistency} \\
The extent to which different predictors trained on the same task do not exhibit logical contradictions.
\label{def:consistency}
\end{definition}

A statement, or a set of statements, is said to be logically consistent when it has no logical contradictions.
A logical contradiction occurs when both a statement and its negation are found to be true.
In logic, a fundamental law -- the law of non-contradiction -- is that a statement and its negation cannot both be true simultaneously.  
Similarly, we measure the consistency between explanations by ensuring that contradictory predictions lead to different explanations.

Following this definition, if the same explanation gets associated with two contradictory predictions the explanation is said to be inconsistent. 
This means avoiding the case where for an observation $\vx \in \V_i$, two predictors $\f_i, \f_j$ (where $i \neq j$), trained on the same task, give the same explanation but different predictions:
\vspace{-1.5mm}\begin{align}
    \label{eq:consistency}
    \f_i(\vx) \neq \f_j(\vx) \implies \g_{\vx}^{(i)} \neq \g_{\vx}^{(j)}
\end{align}

Nevertheless, we have to define what it means for two explanations to be different. For this, we use a measure of dissimilarity between explanations and a threshold to judge whether the explanations are consistent or not. This threshold will be relative to the distance between explanations when predictions are not contrary.
By measuring the rate of inconsistent explanations, we hope to capture the notion of \consistency~for explanations.

 \vspace{-1.mm}\subsection{$k$-Fold Cross-Training}
\label{methodology}
\begin{figure*}[ht]
  \includegraphics[width=0.99\textwidth]{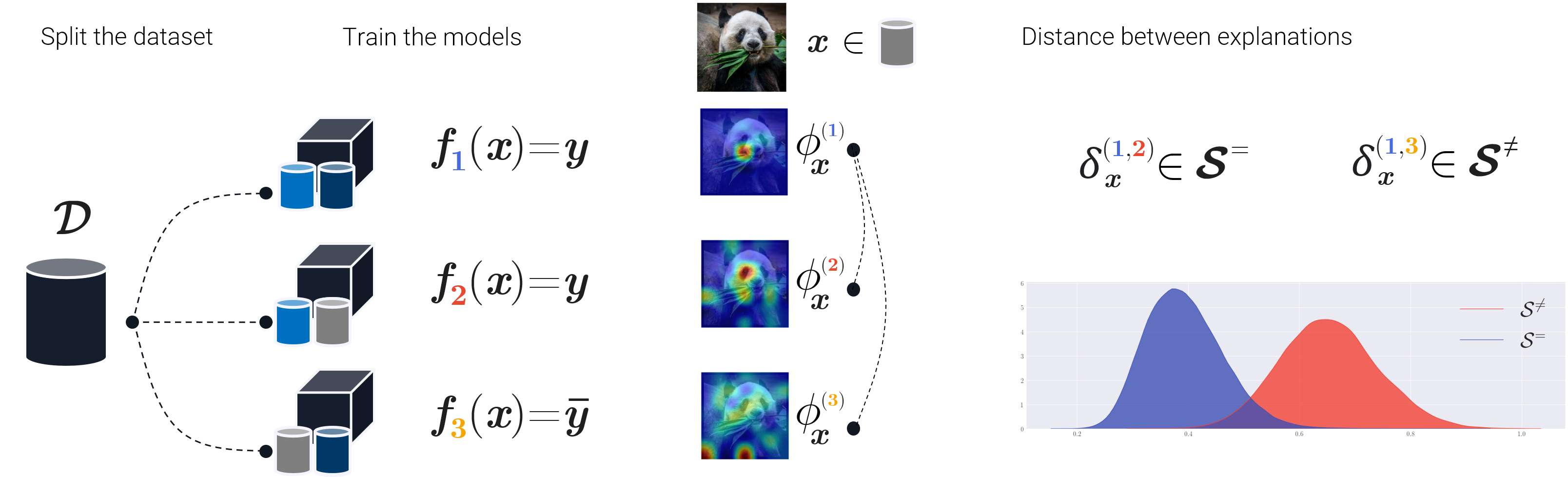}
  \caption{Application of the proposed procedure for $3$ \textit{folds}. Each predictor is trained on two of the 3 \textit{folds}, e.g, $\f_{\textcolor{blue}{1}}$ is trained on $\D \setminus \V_1$. For a given sample $\vx$ such that $\vx \in \V_1$, the explanations for each predictors are calculated ($\phi_{\vx}^{(\textcolor{blue}{1})}, \phi_{\vx}^{(\textcolor{red}{2})}, \phi_{\vx}^{(\textcolor{yellow}{3})}$). The distance between $\phi_{\vx}^{(\textcolor{blue}{1})}$ and the  other two explanations  $\phi_{\vx}^{(\textcolor{red}{2})}, \phi_{\vx}^{(\textcolor{yellow}{3})}$ are computed. 
  All distances for which predictions do not contradict each other are added to $\Sm$ while the others are added to $\Sp$ (note that this is the case for $\delta^{(\textcolor{blue}{1},\textcolor{yellow}{3})}_{\vx}$ since $\f_{\textcolor{blue}{1}}(\vx) \neq \f_{\textcolor{yellow}{3}}(\vx)$).
  }
  \label{fig:methodology}
  \vspace{-3mm}
\end{figure*}

We recall that our data set is divided into $k$-\textit{folds} of the same size $\D = \{\V_i\}_{i=0}^k$, and that each predictor is trained through a learning algorithm $\f_i = \A(\V \setminus \V_i)$.
We assume that the predictors exhibit comparable accuracies across folds. In our experiments, we ensure a similar accuracy on the test set.

We will now measure the distances between two explanations associated with these different predictors.
To be more precise, we are really only interested in computing $\dg$ (see Eq.~\ref{eq:quantity_of_interest}):, the distance between two explanations whereby one of the two predictors was not fitted on $\vx$. Otherwise, it may be trivial for two predictors that were trained on that sample to yield the same explanation -- especially if overfitting occurs (see Fig. \ref{fig:methodology}).

In the case where both predictors gave a correct prediction, a small distance between the two explanations suggests that the explanations receive support from several samples. In other words, the fact that  explanations do not vary widely when adding or removing a particular sample or set of samples suggests good \representativity. Alternatively, if  the two predictors give contrary predictions, the corresponding explanations should be different. Indeed, the very notion of \consistency~between explanations implies that the same explanation cannot account for two different outcomes. 

We separate distances into two sets, $\Sm$~when the predictors have made correct predictions s.t. it is desirable to have a small distance between explanations, $\Sp$~when one of the predictors have given a wrong prediction s.t. it is desirable to have higher distances between the pairs of explanations. The case where both predictors give a bad prediction is ignored (for details, see the Alg.~\ref{alg:procedure} in the appendix).

\vspace{-3.0mm}\begin{align}
    \Sm &= \{ \dg : \f_i(\vx) = \vy \land  \f_j(\vx) = \vy \} \\
    \Sp &= \{ \dg : \f_i(\vx) = \vy \oplus  \f_j(\vx) = \vy \}
\end{align}
$$ \hspace{10000pt minus 1fil} \forall (i, j) \in \{1, ..., k\}^2 ~s.t.~ i \neq j, ~\forall (\vx, \vy) \in \V_i  \hfilneg $$

\subsection{Mean generalizability : \mR}
\label{MeGemeasure}
From Def.~\ref{def:representativity}, the distance between explanations arising from predictors trained on a dataset that contained vs. did not contain a given sample should be small. 
As those distances are contained in $\Sm$, one way to measure the \representativity~of explanations is to compute the average distance over $\Sm$.

As a reminder, the average of $\Sm$ corresponds to the average change of explanation when the sample is removed from the training set. This change is related to the \representativity~of the explanation: the more representative an explanation is, the more it persists when we remove a point.

To ensure a high value for low distances, we define the \mR~measure as a similarity measure:
\begin{align} 
    MeGe &= \Big(1 + \frac{1}{|\Sm|}\sum_{\delta ~\in~ \Sm} \delta \Big)^{-1}
    \label{eq:mege}
\end{align}

Explanations with good \representativity~ will therefore be associated with higher similarity scores between explanations (close to 1).

\vspace{-1mm}\subsection{Relative consistency : \mC}
\label{ReComeasure}
\newcommand{\thr}{\gamma}
\newcommand{\elem}{\delta}

From Def.~\ref{def:consistency} and Eq.~\ref{eq:consistency}, explanations arising from different predictors are said to be consistent if they are close when the predictions agree with one another.
As a reminder, the distance between explanations for the consistent predictions are represented by $\Sm$, and those associated with inconsistent predictions by $\Sp$. 
Visually, we seek to maximize the shift between the corresponding distributions for the sets $\Sm$ and $\Sp$.
Formally, we are looking for a distance value that separates $\Sm$ and $\Sp$, e.g., such that all the lower distances belong to $\Sm$ and the higher ones to $\Sp$. The clearer the separation, the more consistent the explanations are.
In order to find this separation, we introduce \mC, a statistical measure based on maximizing the balanced accuracy.

Where $ \Sa = \Sm \cup \Sp $ and $\thr \in \Sa$ a fixed threshold value, we can define the true positive rate $TPR$ as the rate for which distances below a threshold come from consistent predictions among all distances below the threshold $TPR(\thr) = \frac{|\{\elem \in \Sm : \elem \leqslant \thr \}|} {|\{ \elem \in \Sa~:~ \elem \leqslant \thr \}| \hfill}$. In a similar way, $TNR$ denotes the rate for which distances above a threshold come from opposite predictions among all the distances above the threshold $TNR(\thr) = \frac{|\{ \elem \in \Sp : \elem > \thr \}|}{|\{ \elem \in \Sa~:~ \elem > \thr \}| \hfill}$. Basing our measure on these rates allows us to assess the quality of these explanations independently of the accuracy of the predictor, we define \mC~ as the maximal balanced accuracy:

\vspace{-2mm}\begin{align}
    ReCo &= \max_{\thr \in \Sa}\ TPR(\thr) + TNR(\thr) - 1, 
    \label{eq:reco}
\end{align}

with a score of 1 indicating perfect consistency of the predictors' explanations, and a score of 0 indicating a complete inconsistency.
    
\vspace{-2mm}\section{Experiments}
\label{experiments}
We carried out three sets of experiments using a variety of neural network architectures and explanation methods. 
The first one consisted in ensuring the functioning and the reliability of the measures \mC~and \mR~via a simple sanity check done over a large number of predictors ($175$ in total).
The second set of experiments consisted in highlighting a limitation of the fidelity measure -- namely its independence with respect to the quality of the explanations. This underlines the need for new measures that are dedicated to explanations and not to methods. We developed these considerations in a dedicated section where we demonstrate an application to the selection of a method using the two new criteria \mR~and \mC.
Finally, in a third set of experiments, we showed quantitatively that some predictors are more interpretable: our analyses revealed that 1-Lipschitz neural networks yield explanations that are more coherent and representative.

\subsection{Setup}

For all experiments, we used 5 splits ($k = 5$), i.e., $5$ predictors with comparable accuracy ($ \pm 3\%$), which allows us to study the explanations in common training conditions (80\% of the data are used for training and 20\% for testing).

For ILSVRC 2012, our predictors are based on a ResNet-50 architecture~\cite{he2016deep}, and a ResNet-18 for the other datasets (see appendix~\ref{ap:models} for details on each predictor).

\vspace{-3.0mm}\paragraph{Explanation methods}

In order to produce the necessary explanations for the experiment, we used $7$ methods of explanation. 
The methods selected are those commonly found in the literature in addition to one control method (Random).

The explanations methods chosen are as follow: Saliency \textbf{(SA)}~\cite{simonyan2013deep}, Gradient $\odot$ Input \textbf{(GI)}~\cite{ancona2017better}, Integrated Gradients \textbf{(IG)}~\cite{sundararajan2017axiomatic}, SmoothGrad \textbf{(SG)}~\cite{smilkov2017smoothgrad}, Grad-CAM \textbf{(GC)}~\cite{Selvaraju_2019}, Grad-CAM++ \textbf{(G+)}~\cite{chattopadhay2018grad} and RISE \textbf{(RI)}~\cite{petsiuk2018rise}. Further information on these methods can be found in the appendix~\ref{ap:methods}.

\vspace{-3.0mm}\paragraph{Datasets}
\label{expDataset}

We applied the procedure described above and evaluated the proposed measures for each of the degradations on $4$ image classification datasets: 

\textbf{ILSVRC 2012}~\cite{imagenet_cvpr09}: a subset of the ImageNet dataset from which we randomly selected $50$ classes. The size of the images considered was $224 \times 224$.
The reduced number of classes being sufficient to show that the metrics pass the test performed in \ref{ApDegradation} even in the case of high dimensional images.

\textbf{CIFAR10}~\cite{krizhevsky2009learning}: a low-resolution labeled datasets with 10 classes respectively, consisting of $60,000$ ($32 \times 32$) color images. 

\textbf{EuroSAT}~\cite{helber2019eurosat}: a labeled dataset with $10$ classes consisting of $27,000$ color images ($64 \times 64$) from the Sentinel-2 satellite.

\textbf{Fashion MNIST}~\cite{xiao2017fashion}: a dataset containing $70,000$ low-resolution ($28 \times 28$) grayscale images labeled in $10$ categories.

\begin{figure*}[t]
    \centering
    \includegraphics[width=0.98\textwidth]{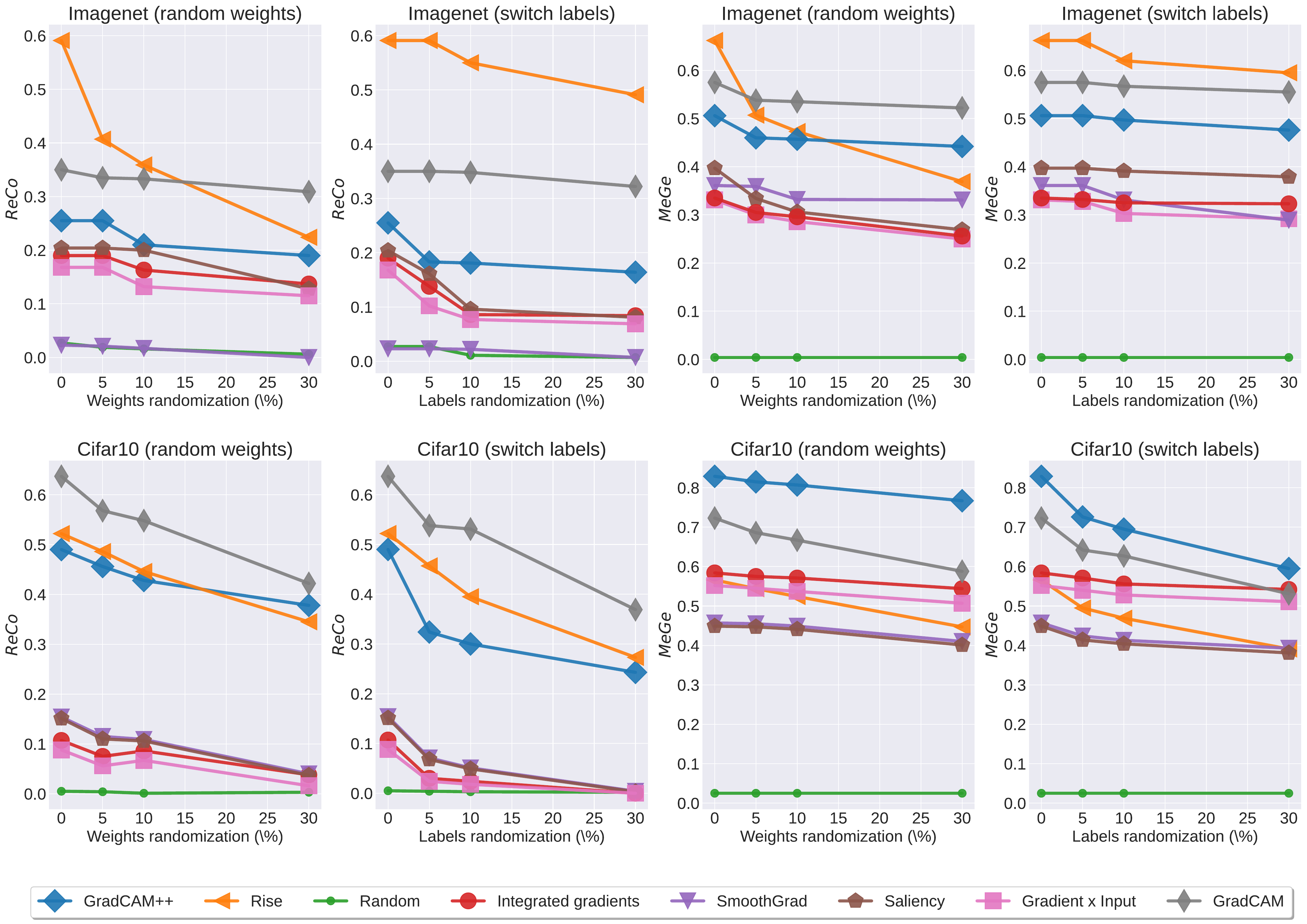}
    \caption{\textbf{\mR~ and \mC~scores} for predictors trained with no degradations (first point from the left), as well as for progressively randomized predictors and predictors trained with switched labels.
    For all the methods tested, the more the predictor is degraded, the more the \consistency ~and \representativity~scores drop, which means that the associated metrics pass the sanity check.
    \textbf{Top} ImageNet. \textbf{Bottom} Cifar-10.
    }
    \label{fig:sanity_check}
    \vspace{-3mm}
\end{figure*}

\vspace{-4mm}\paragraph{Distance over explanations}
\label{consideredDistances}

The procedure introduced in section~\ref{methodology} requires to define a distance between two explanations derived for the same sample. 
Since a feature attribution consists of ranking the features most sensitive to the predictor's decision, it seems natural to consider the  Spearman rank correlation~\cite{spearman1904measure} to compare the similarity between explanations. Several authors have provided theoretical and experimental arguments in line with this choice~\cite{ghorbani2017interpretation, adebayo2018sanity, tomsett2019sanity}. However, it is important to note that the problem of measuring similarity between explanations is still an open problem. We conduct two sanity checks: spatial correlation, and noise test on several candidates distances to ensure they could respond to the problem.  
The distances tested were built from: 1-Wasserstein distance (the Earth mover distance from~\cite{flamary2017pot}), Sørensen–Dice~\cite{dice1945} coefficient, Spearman rank correlation, SSIM~\cite{ssim2004}, and $\ell_1$ and $\ell_2$ norms.
In line with prior work, we chose to use one minus the absolute value of the Spearman rank correlation (see~\ref{ap:distances} for more details).

\subsection{Sanity check for explanation measures}
\label{ApDegradation}

Our first set of experiments aims to ensure that the propose metrics approximate the desired quantities by performing a sanity check: on average, as the learning is degraded, we expect to see an overall increase in the number of specific and inconsistent explanations.
To ensure that the metric captures these notions, we applied two different types of degradation on the predictors for each data set: weight randomizations and label shuffling.
\begin{itemize}
\setlength\itemsep{-0.2em}
\item Randomizing the weights, inspired by~\cite{adebayo2018sanity}. We gradually randomize $5$\%, $10$\% and $30$\% of the predictor layers by adding Gaussian noise. By degrading the weights learned by the network, we expect to find degraded explanations. 
\item Shuffling of labels, inspired by~\cite{neyshabur2017exploring, adebayo2018sanity} the predictors are trained on a data set with $5$\%, $10$\% and $30$\% of bad labels. By artificially breaking the relationship between the labels, we expect the explanations to lose their consistency.
\end{itemize}

The \mR~measure encodes the \representativity~of the explanations, which is related to the ability of the predictor to derive general strategies. 
Thus, the degradation of the parameters of a predictor directly affects these strategies.
Fig.~\ref{fig:sanity_check} shows the correlation of the measures with the intensity of the degradation applied: \mR~and \mC~capture the degradation of the explanation and pass the sanity check.

We note that all the tested methods perform better than the random baseline (random). However, the drop in score, is not the same and some methods are more sensitive to predictor changes, such as Grad-CAM or RISE, in accordance with previous work~\cite{adebayo2018sanity, sixt2020explanations}. 
It was subsequently observed that this sensitivity seems to translate into a better \fidelity~score for the methods.
Nevertheless, it should be noted that this sanity test is a necessary but not sufficient condition for a \representativity~and \consistency~metric.

\subsection{The implications of the fidelity metric}

To mark the difference between the proposed measures and the \fidelity, we applied the $\mu F$ measure from~\cite{aggregating2020} (see supplementary document~\ref{ap:fidelity}) to the normally trained predictors and those progressively degraded. We seek to verify that this property does not pass the sanity check: the fidelity measure is invariant to the performancee of the predictor as well as to the quality of its explanations.
For $\mu F$, the score obtained is averaged over $10,000$ test samples, with $0$ for baseline. The size of the $|S|$ subset is $15$\% of the image.

\begin{figure}[ht]
    \centering
    \includegraphics[width=0.48\textwidth]{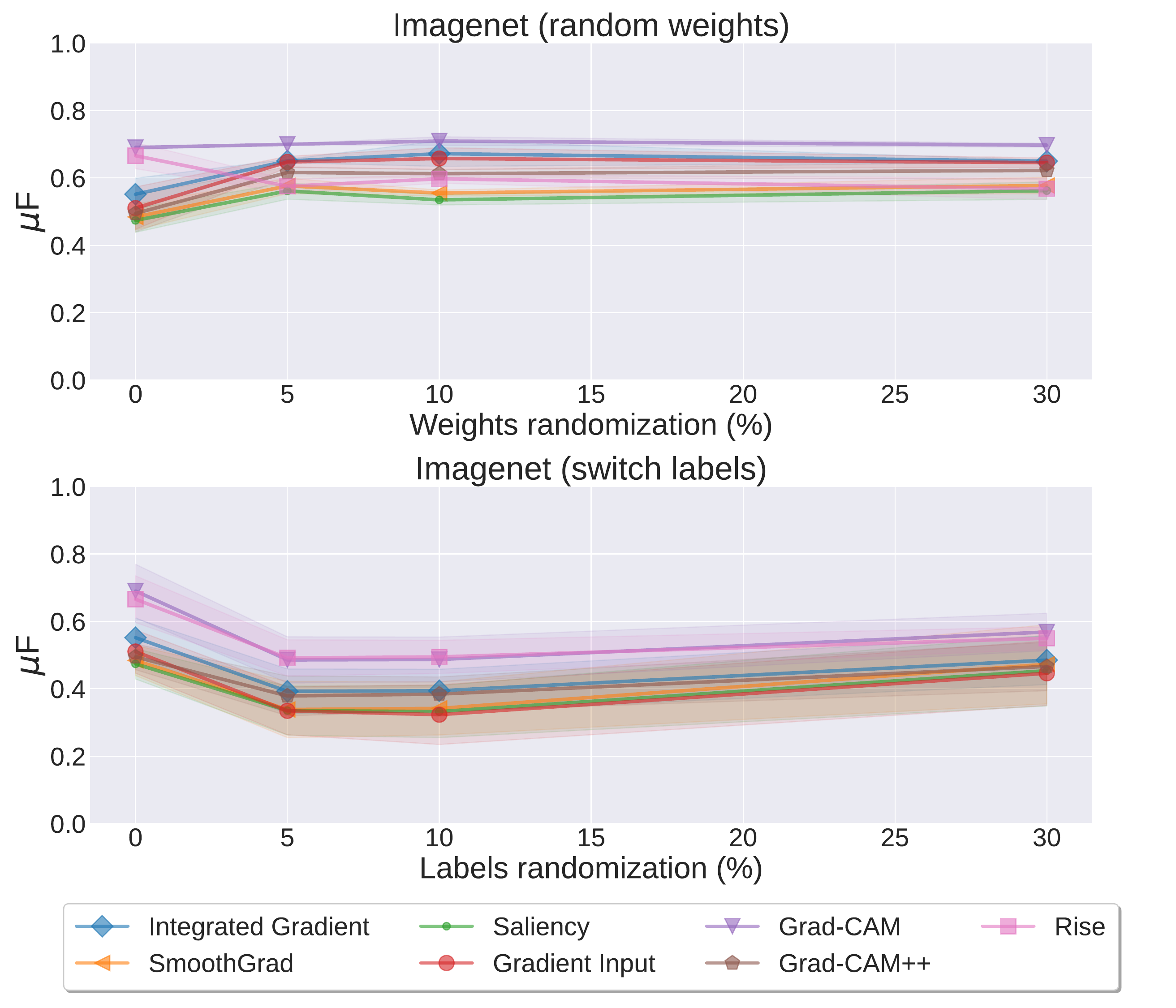}
    \caption{
    \fidelity~scores (Equation~\ref{eq:fidelity}) on ImageNet for normally trained ResNet-50 predictors (first point on the left) as well as for progressively randomized predictors and predictors trained with switched labels.
    Even a strong degradation of the predictor does not impact the \fidelity~of the tested methods. Hence, the \fidelity~is intended to ensure that the explanations correctly reflect the underlying strategies of the model, regardless of whether these strategies are general or consistent.
    }
    \label{fig:fidelity}
\end{figure}

As shown in Fig.~\ref{fig:fidelity}, predictor degradation does not impact the \fidelity~ metric on the methods tested. 
The \fidelity~property is essential in a good explanation since it allows us to make sure that we are studying the strategies of the predictor.
However, it is not sufficient: if the explanation reflects well the strategies of the predictor, the latter may use specific and inconsistent strategies. In that, the \fidelity~measure is only a first step towards a good explanation.

\subsection{Method selection criterion}

The \mR~and \mC~measures can be used as additional criteria for choosing an explainability method.  
As a reminder, a good method should provide explanations that are as faithful as possible and, if possible, consistent and representative.
Thus, the tested methods can be compared using the scores obtained for these measures. We note that these measures are complementary in that the fidelity score can be interpreted as a confidence bound on the other measures performed on the explanations.

\begin{table}[h]
    \centering
    \scalebox{0.85}{
        \begin{tabular}{l lllllll}
        \toprule
        \textbf{ImageNet} & SA & GI & IG & SG & GC & G+ & RI \\
        \midrule
        $\mu F$ & 0.47 & 0.51 & 0.55 & 0.48 & \textbf{0.69} & 0.49 & \underline{0.67} \\
        \mR     & 0.40 & 0.50 & \underline{0.58} & 0.36 & 0.34 & 0.33 & \textbf{0.66} \\
        \mC     & 0.20 & 0.17 & 0.16 & 0.02 & \underline{0.35} & 0.26 & \textbf{0.59} \\
        \bottomrule \\
        \end{tabular}
    }
    \caption{\consistency, \representativity~and \fidelity~score for ResNet-50 models on ImageNet. Higher is better. The first and second best results are respectively in \textbf{bold} and \underline{underlined}.}
    \label{tab:metrics_imagenet}
\end{table}

Table~\ref{tab:metrics_imagenet} reports the \fidelity~($\mu F$), \consistency~(\mC) and \representativity~(\mR) scores obtained for the ResNet-50 predictors trained without degradation on ImageNet. 
We can exploit a selection criterion from the differences in scores.
First of all, we notice that the two methods obtaining a good fidelity score are RISE and Grad-CAM, they reflect well the predictor functioning. 
Their high fidelity score acts as a confidence bound on the \mR~and \mC~metrics: by correctly transcribing the functioning of the predictor, we obtain at the same time the \representativity~and the \consistency~of the explanations.
This score can then be used as a criterion to separate RISE from Grad-CAM. In view of the differences between the \mR~and \mC~scores, RISE method seems preferable.

Concerning the \representativity~score, it is important to note that two methods tested here involve the element-wise product of the explanation with the input: Integrated Gradients and Gradient Input. This operation could eliminate the attribution score on a part of the image, thus reducing the distance between the two explanations. The result is a better \mR~score which is in fact due to the dominance of input in the element-wise product.

It can be observed that the change of predictor has an effect on this ranking, and that a good method of explainability must be chosen according to a context: predictor and data set. However, even considering these effects, the experiments carried out suggest $3$ methods that give faithful, representative and consistent explanations: Grad-CAM, Grad-CAM++ and RISE (for more results on Cifar-10, EuroSAT and Fashion MNIST, see appendix~\ref{ap:results}).

\subsection{Towards predictors with better explanations}

In an attempt to find predictors that give better explanations, we extend the experience on the Cifar-10 dataset by adding a family of 1-Lipschitz networks. Indeed different works mention the Lipschitz constrained networks as particularly robust~\cite{usama2018robust, scaman2019lipschitz, pauli2020training, louislip} and have good generalizability. As a reminder, a $\f$ function is called $L$-Lipschitz, with $L \in \mathbb{R}^+$ if 
$| \f(\vx_1) - \f(\vx_2) | \leq L |\vx_1 - \vx_2|$
For every pair $(\vx_1, \vx_2) \in \X^2$. The smallest of these $L$ is called the Lipschitz constant of $\f$. This constant certifies that the gradients of the function represented by the deep neural network are bounded  (given a norm) and that this bound is known. This robustness certificate also comes with new generalization bounds that critically rely on the Lipschitz constant of the neural network~\cite{von2004distance, neyshabur2017exploring, bartlett2017spectrallynormalized}.

The predictors were trained using the Deel-Lip library~\cite{deelLip}. All the predictors, including the 1-Lipschitz, have comparable accuracy ($78 \pm 4\%$). To our knowledge, no previous work has made the link between Lipschitz networks and the chosen explainability methods.

\begin{figure}[ht]
    \centering
    \includegraphics[width=0.5\textwidth]{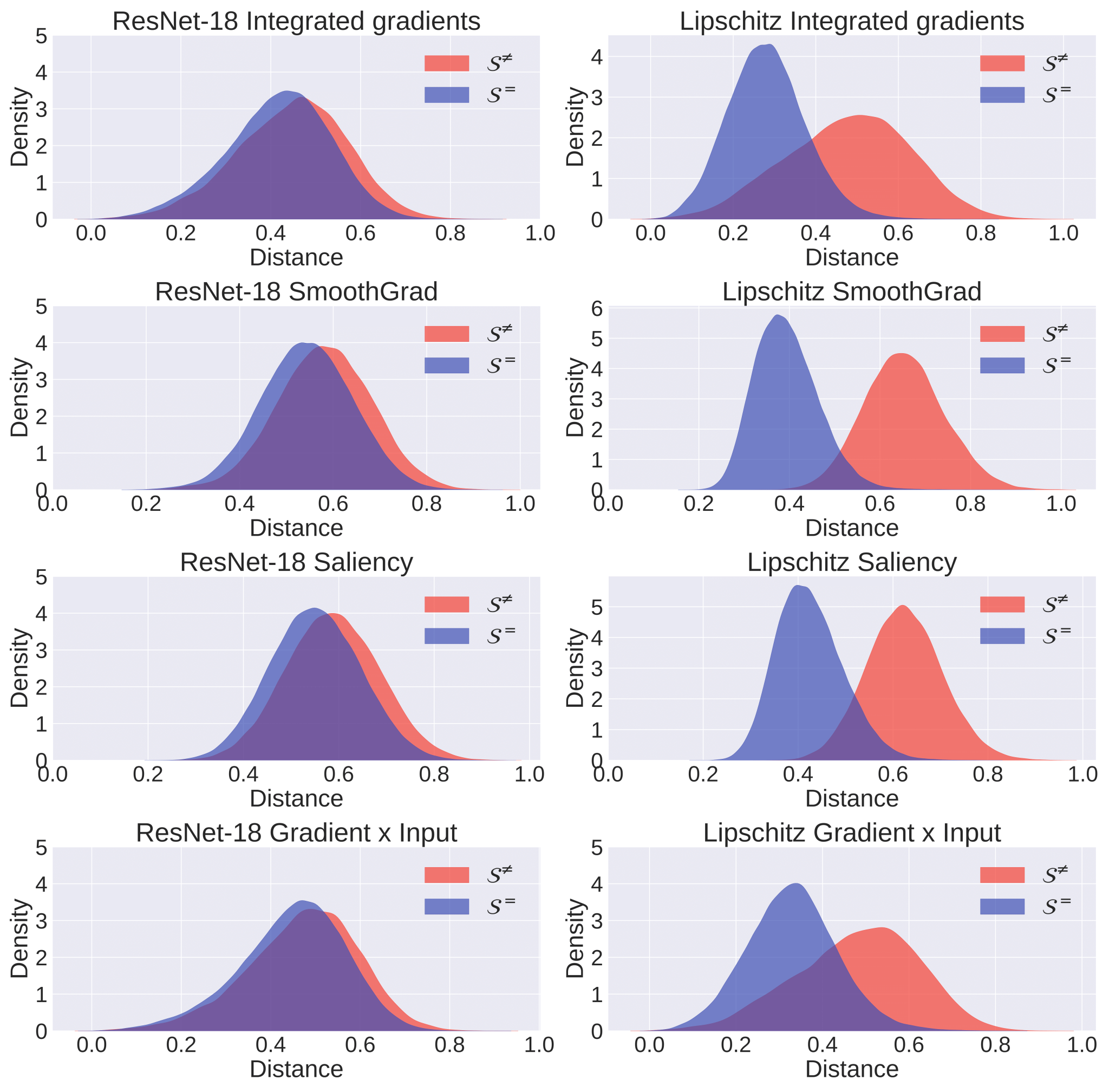}
    \caption{Lipschitz predictors (right column) on Cifar10.
    As explained in this paper, a clear separation between the $\Sm$ and $\Sp$ histograms is a sign of consistent explanations.
    }
    \label{fig:lipVsNormalDistrib}
    \vspace{-3mm}
\end{figure}

The Fig.~\ref{fig:lipVsNormalDistrib} shows the difference in $\Sp$ and $\Sm$ between ResNet and 1-Lipschitz predictors. In the left column, the results come from ResNet-18 predictors trained on Cifar-10 while the right column is dedicated to 1-Lipschitz predictors. We observe a clear improvement of the consistency and generalization of the explanations respectively as a result of better separation of the histograms and a smaller expectation of $\Sm$. SmoothGrad is the method that obtains the most consistent explanations as indicated in the table~\ref{tab:lip_reco}, in front of RISE and Saliency (more results in the supplementary material~\ref{ap:results} Fig.~\ref{fig:lipVsNormalDistrib2}). 

\begin{table}[h]
    \centering
    \scalebox{0.88}{
        \begin{tabular}{l lllllll}
        \toprule
        \mR & IG & SG & SA & GI & GC & G+ & RI \\
        \midrule
        ResNet-18    & 0.58 & 0.46 & 0.45 & 0.55 & 0.72 & \textbf{0.83} & 0.57 \\
        1-Lipschitz  & \textbf{0.72} & \textbf{0.60} & \textbf{0.58} & \textbf{0.67} & \textbf{0.75} & 0.54 & \textbf{0.85} \\
        \bottomrule \\
        \end{tabular}
    }
    \caption{\mR~scores obtained by 1-Lipschitz models and ResNet-18 models on Cifar10. Higher is better. For almost all methods, the \representativity~of explanations increases significantly on 1-Lipschitz models.}
    \label{tab:lip_mege}
     \vspace{-1mm}
\end{table}

Concerning \mR, the results reported in Table~\ref{tab:lip_mege} show an improvement in the \representativity~of the explanations for the 1-Lipschitz predictors. Indeed, the \representativity~score has increased compared to the ResNet predictors for all tested methods, except Grad-CAM++. 

\begin{table}[h]
    \centering
    \scalebox{0.88}{
        \begin{tabular}{l lllllll}
        \toprule
        \mC & IG & SG & SA & GI & GC & G+ & RI \\
        \midrule
        ResNet-18     & 0.11 & 0.15 & 0.15 & 0.09 & 0.64 & \textbf{0.49} & 0.52 \\
        1-Lipschitz   & \textbf{0.60} & \textbf{0.90} & \textbf{0.81} & \textbf{0.50} & \textbf{0.67} & 0.24 & \textbf{0.84} \\
        \bottomrule \\
        \end{tabular}
    }
    \caption{\mC~scores obtained by 1-Lipschitz models and ResNet-18 models on Cifar10. Higher is better. For almost all methods, the \consistency~of explanations increases significantly on 1-Lipschitz models.}
    \label{tab:lip_reco}
     \vspace{-1mm}
\end{table}

Like \mR, the results in Table~\ref{tab:lip_reco} show an improvement for the 1-Lipschitz predictors in the \consistency~of the explanations for all the methods tested except for Grad-CAM++, reflecting the more marked separation between the two histograms of $\Sm$ and $\Sp$ in Fig.~\ref{fig:lipVsNormalDistrib}.

In general, the experiments carried out allow us to observe a clear improvement in the quality of explanations from the 1-Lipschitz predictor.
These encouraging results show that there is a close link between the methods used and predictor architectures, as well as the usefulness of Lipschitz networks for explainability.
Furthermore, it underlines the fact that the search for new methods is not the only path to explainability: the search for predictors with better explanations is another under-exploited avenue. 


\vspace{-2mm}\section{Conclusion}
\label{conclusion}

We introduced a procedure to derive two new measures to characterize important properties of a good explanation: \representativity~and~\consistency.
We highlight the fact that \fidelity~is intended to ensure that the explanations correctly reflect the underlying strategies of the model, regardless of whether these strategies are general or consistent.
Finally, as a case in point, we presented a novel analysis using 1-Lipschitz networks. We  used our measures to quantify the consistency of their explanations and showed that this class of networks gives much more stable and trustworthy explanations compared to standard neural networks. 


We see the present work as constituting a necessary next step in characterizing good explanations -- towards the quest for more explainable ML models.

\vspace{-2mm}\section*{Acknowledgement}
This work was conducted as part of the DEEL project\footnote{https://www.deel.ai/}. Funding was provided by ANR-3IA Artificial and Natural Intelligence Toulouse Institute (ANR-19-PI3A-0004). 
Additional funding to TS was provided by ONR (N00014-19-1-2029), NSF (IIS-1912280 and EAR-1925481), DARPA (D19AC00015), NIH/NINDS (R21 NS 112743). The computing hardware was supported in part by NIH Office of the Director grant S10OD025181.
We thank Melanie Ducoffe and Mikael Capelle of the DEEL team for insightful comments which have helped improved the manuscript.

{\small
\bibliographystyle{ieee_fullname}
\bibliography{egbib}
}
\clearpage
\appendix
\newpage

\section{Method Details}
\begin{algorithm}[h]
\caption{Training procedure to compute $\Sm$ and $\Sp$}
\begin{algorithmic}
\label{alg:procedure}

\REQUIRE $k \in \mathbb{N}_{\geq 2} \ ,\ \D = \{ \V_i \}_{i=1}^k $
\STATE $\Sm \gets{} \{\}$, $\Sp \gets{} \{\}$

\FORALL{$ i \in  \{1,~\ldots{},~k \}  $}
\STATE \textbf{Train} $\f_i$ on $\D \setminus \V_i$
\FORALL{$(\vx, \vy) \in \D$}
\STATE \textit{// generate explanations on all dataset} 
\STATE $ \g_{\vx}^{(i)} \gets{} \G(\f_i, \vx)$
\ENDFOR
\ENDFOR

\FORALL{$ i \in  \{1,~\ldots{},~k \}$}
\FORALL{$(\vx, \vy) \in \V_i$}
\FORALL{$ j \in  \{1,~\ldots{},~k \mid i \neq j \}$}

\STATE \textit{// $\f_{j}$ was trained on $\vx$, $\f_{i}$ was not} 
\STATE $ \delta_{\vx}^{(i,j)} \gets{} d( \g_{\vx}^{(i)}, \g_{\vx}^{(j)} )$

\IF{$\f_{i}(\vx) = \vy $ \AND $\f_{j}(\vx) = \vy$}
\STATE \textit{// both model are correct}
\STATE $\Sm \gets{} \Sm \cup \{ \delta_{\vx}^{(i,j)} \}$

\ELSIF{$\f_i(\vx) = \vy $ \OR $\f_j(\vx) = \vy$}
\STATE \textit{// only one model is correct}
\STATE $\Sp \gets{} \Sp \cup \{ \delta_{\vx}^{(i,j)} \}$

\ENDIF
\ENDFOR
\ENDFOR
\ENDFOR

\STATE \textbf{Return} $\Sm, \Sp$

\end{algorithmic}
\end{algorithm}

\section{Explanation methods}
\label{ap:methods}

In the following section, the formulation of the different methods used is given. As a reminder, we focus on a classification model $\f : \mathbb{R}^d \to \mathbb{R}^C$ where $C$ is the number of classes. 
We assume $\f_c(\vx)$ the logit score (before softmax) for class $c$.
An explanation method provides an attribution $\g \in \mathbb{R}^d$ for each input feature from a model and an input of interest. Each value then corresponds to the importance of this feature for the model results.

\textbf{Saliency Map (SA)} is a visualization techniques based on the gradient of a class score relative to the input, indicating in an infinitesimal neighborhood, which pixels must be modified to most affect the score of the class of interest.

$$ \G^{SA}(\vx) = \Big\lvert\frac{\partial \f_c(\vx)}{\partial \vx}\Big\rvert $$

\textbf{Gradient $\odot$ Input (GI)} is based on the gradient of a class score relative to the input, element-wise with the input, it was introduced to improve the sharpness of the attribution maps. A theoretical analysis conducted by \cite{ancona2017better} showed that Gradient $\odot$ Input is equivalent to $\epsilon$-LRP and DeepLIFT methods under certain conditions: using a baseline of zero, and with all biases to zero.

$$ \G^{GI}(\vx) = \vx \odot \Big\lvert\frac{\partial \f_c(\vx)}{\partial \vx}\Big\rvert $$

\textbf{Integrated Gradients (IG)} consists of summing the gradient values along the path from a baseline state to the current value. The baseline is defined by the user and often chosen to be zero. This integral can be approximated with a set of $m$ points at regular intervals between the baseline and the point of interest. In order to approximate from a finite number of steps, we use a Trapezoidal rule and not a left-Riemann summation, which allows for more accurate results and improved performance (see \cite{sotoudeh2019computing} for a comparison). The final result depends on both the choice of the baseline $\vx_0$ and the number of points to estimate the integral. In the context of these experiments, we use zero as the baseline and $m = 60$.

$$ \G^{IG}(\vx) = (\vx - \vx_0) \int_0^1 \frac{\partial \f_c( \vx_0 + \alpha(\vx - \vx_0) ) }{\partial \vx} d\alpha $$

\textbf{SmoothGrad (SG)} is also a gradient-based explanation method, which, as the name suggests, averages the gradient at several points corresponding to small perturbations (drawn i.i.d from a normal distribution of standard deviation $\sigma$) around the point of interest. The smoothing effect induced by the average help reducing the visual noise, and hence improve the explanations. In practice, Smoothgrad is obtained by averaging after sampling $m$ points. In the context of these experiments, we took $m = 60$ and $\sigma = 0.2$ as suggested in the original paper.

$$ \G^{SG}(\vx) = \underset{\varepsilon ~\sim~ \mathcal{N}(0, \bm{I}\sigma^2)}{\mathbb{E}}\Big{[}\frac {\partial{\f_c(\vx + \varepsilon)} }{ \partial{\vx} }\Big{]}  $$

\textbf{Grad-CAM (GC)} can be used on Convolutional Neural Network (CNN), it uses the gradient and the feature maps $\bm{A}^{(k)}$ of the last convolution layer. More precisely, to obtain the localization map for a class, we need to compute the weights $\alpha_c^{(k)}$ associated to each of the feature map activation $\bm{A}^{(k)}$, with $k$ the number of filters and $Z$ the number of features in each feature map we define $\alpha_c^{(k)} = \frac{1}{Z} \sum_i\sum_j \frac{\partial{\f_c(\vx)}}{\partial A^{(k)}_{ij}} $ and $$\G^{GC} = max(0, \sum_k \alpha_c^{(k)} \bm{A}^{(k)}) $$
Notice that the size of the explanation depends on the size (height, width) of the last feature map, a bilinear interpolation is performed in order to find the same dimensions as the input.

\textbf{Grad-CAM++ (G+)} is an extension of Grad-CAM combining the
positive partial derivatives of feature maps of a convolutional layer with a weighted special class score. The weights $\alpha_c^{(k)}$ associated to each feature map is computed as follow : 

$$\alpha_k^c = 
    \sum_i \sum_j [
    \frac{ \frac{\partial^2 \f_c(\vx) }{ (\partial A_{ij}^{(k)})^2 } }
    { 2 \frac{\partial^2 \f_c(\vx) }{ (\partial A_{ij}^{(k)})^2 } + \sum_i \sum_j A^{(k)}_{ij}  \frac{\partial^3 \f_c(\vx) }{ (\partial A_{ij}^{(k)})^3 } }
    ]
$$ 

\textbf{RISE (RI)} is a black-box method that consist of probing the model with randomly masked versions of the input image to deduce the importance of each pixel using the corresponding outputs. The binary masks $\bm{m} \sim \mathcal{M}$ are generated in a subspace of the input space, then upsampled with a bilinear interpolation (once upsampled the masks are no longer binary).

For ImageNet the number of masks was $m = 4000$, for all the other datasets $m = 1000$.

$$ \G^{RI}(\vx) = \frac{1}{\mathbb{E}(\mathcal{M}) N} \sum_{i=0}^N \f_c(\vx \odot \bm{m_i}) \bm{m_i} $$

\section{Fidelity}
\label{ap:fidelity}

Various fidelity metrics have been proposed that essentially measure the correlation between input variables and the drop in score when these variables are set to a baseline state \cite{samek2015evaluating, yeh2019infidelity, rieger2020irof, petsiuk2018rise}. 
In this work, we use $\mu F$ from \cite{aggregating2020}: 

\begin{equation}
\label{eq:fidelity}
\mu F = \underset{S \subseteq \{1, ..., d\} \atop |S| = k}{\operatorname{Corr}}\left( \sum_{i \in S} \G(\f, \vx)_i  , \f(\vx) - \f(\vx_{[\vx_i = \bar{\vx_i}, i \in S]})\right)
\end{equation}
Where $\f$ is a predictor, $\G$ an explanation function, $S$ a subset indices of $\vx$ and $\bar{\vx}$ a baseline reference. The choice of a proper baseline is still an active area of research \cite{sturmfels2020visualizing}.

\section{Considered measures for ReCo}
\label{ApReCo}
As mentioned in when introducing \mC, one would be tempted to use directly a distance between distributions, we briefly explain why we did not make this choice. In addition, we detail an alternative measure, also based on balanced accuracy, which gives consistent results.

A first intuition to measure the shift between the $\Sm$ and $\Sp$ histograms would be to consider the usual measures, such as Kullback-Leibler ($KL$) divergence.

However, these distances are problematic in that the order of the distributions actually matters more than the distance between them, and these two measures can give a good score even when the explanations are inconsistent 
Similarly, considering the 1-Wasserstein measure, we could construct an inconsistent case by exploiting the invariance to the direction of transport.
For these reasons, we have therefore chosen a classification measure, based on maximizing balanced accuracy. Nevertheless, one could also (observing similar results) use the area under the curve (AUC) of the balanced accuracy, such as : 
$$ \mC_{AUC} = \frac{1}{|\Sa|}\sum_{\gamma \in \Sa} TPR(\gamma) + TNR(\gamma) - 1  $$

\section{Models}
\label{ap:models}
As mentioned in the paper, the models used are all (with the exception of 1-Lipschitz networks) ResNet-18, with variations in size and number of filters used. Preserving the increase of filters at each depth by the original factor (x2), we took care to define for each dataset, a base filters value, as the number of filters for the first convolution layer. Another difference concerns the dropout rates used, indeed we had dropout to improve the performance of the tested models. Moreover, it should be remembered that there is no difference in architecture between the normally trained models and the degraded models.

We report here the architecture of the models for each of the datasets:\\
\textbf{Fashion-MNIST} base filters $26$, Dropout $0.4$ ($92$\%, $\pm 1$\%)\\
\textbf{EuroSAT}       base filters $46$, Dropout $0.25$ ($95$\%, $\pm 1$\%)\\
\textbf{Cifar10}      base filters $32$, Dropout $0.25$ ($78$\%, $\pm 4$\%)\\
\textbf{ImageNet}     ResNet50 ($88$\%, $\pm 3$\%)\\

\subsection{Lipschitz models}
The 1-Lipschitz models use spectral regularization on the Dense and Convolutions layers. The architecture is as described in Table \ref{ArchiLipschitz}. 

\begin{table}
    \centering
    \caption{$1$-Lipschitz model architecture for Cifar10.}
    \label{ArchiLipschitz}
    \begin{tabular}{l}
    \midrule
    Conv2D($48$) \\
    PReLU \\
    AvgPooling2D(($2$, $2$)) \\ 
    Dropout($0.2$) \\
    Conv2D($96$) \\ 
    PReLU \\
    AvgPooling2D(($2$, $2$)) \\ 
    Dropout($0.2$) \\
    Conv2D($96$) \\
    AvgPooling2D(($2, 2$)) \\ 
    Flatten \\
    Dense($10$) \\
    \bottomrule
    \end{tabular}
\end{table}

\subsection{Randomization test}

For the randomisation of the model weights, we added noise drawn from a normal distribution $\varepsilon \sim \mathcal{N}(0, 0.5)$ to each convolution layer, with the intensity of the degradation impacting on the number of parameters affected by this noise.

\section{Distances tests}
\label{ap:distances}

\subsection{Spatial correlation}
\begin{figure}[h]
  \centering
  \includegraphics[width=0.48\textwidth]{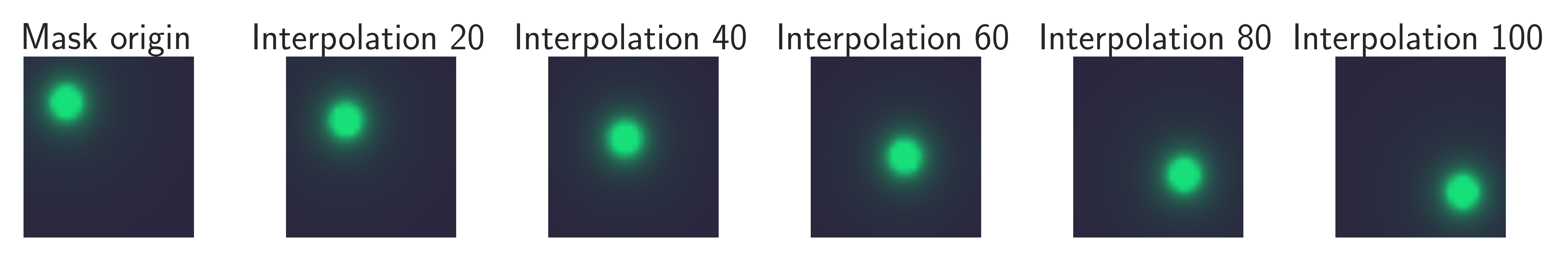}
  \includegraphics[width=0.48\textwidth]{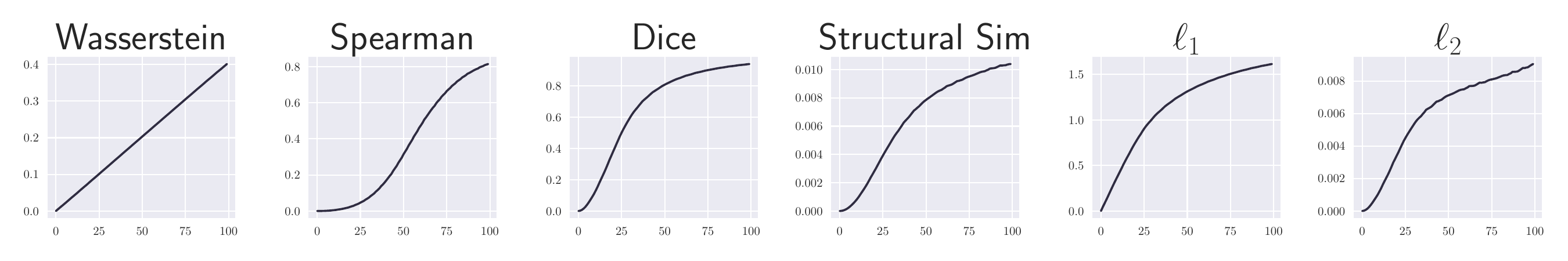}
  \caption{
    Distances with moving interest point. The first line shows the successive interpolations between the baseline image (left), and the target image (right). The second line shows the evolution of the distance between each interpolation and the baseline image.
    }
  \label{dist:move}
\end{figure}

The first test concerns the spatial distance between two areas of interest for an explanation. It is desired that the spatial distance between areas of interest be expressed by the distance used. As a results, two different but spatially close explanations should have a low distance. The test consists in generating several masks representing a point of interest, starting from a left corner of an image of size (32 x 32) and moving towards the right corner by interpolating 100 different masks. The distance between the first image and each interpolation is then measured (see Fig. \ref{dist:move}). 

The different distances evaluated pass this sanity check, i.e. a monotonous growth of the distance, image of the spatial distance of the two points of interest. 

\subsection{Noise test}

The second test concerns the progressive addition of noise. It is desired that the progressive addition of noise to an original image will affect the distance between the original noise-free image and the noisy image. Formally, with $x$ the original image, and $\varepsilon\ \sim\ \mathcal{N}(0, \bm{I}\sigma^2)$ an isotropic Gaussian noise, we wish the distance $d$ to show a monotonic positive correlation $\operatorname{corr}( dist(x, x + \varepsilon), \varepsilon )$.

In order to validate this prerogative, a Gaussian noise with a progressive intensity $\sigma$ is added to an original image, and the distance between each of the noisy images and the original image is measured. For each value of $\sigma$ the operation is repeated 50 times.

\begin{figure}[h]
  \centering
  \includegraphics[width=0.48\textwidth]{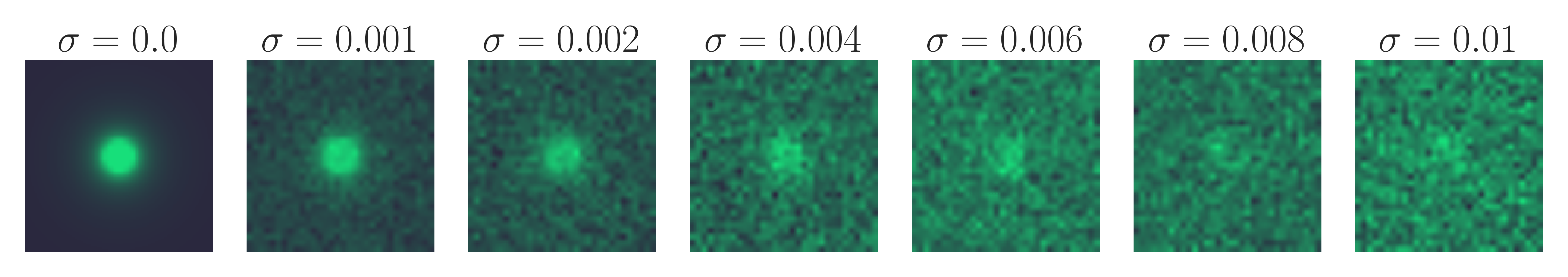}
  \includegraphics[width=0.48\textwidth]{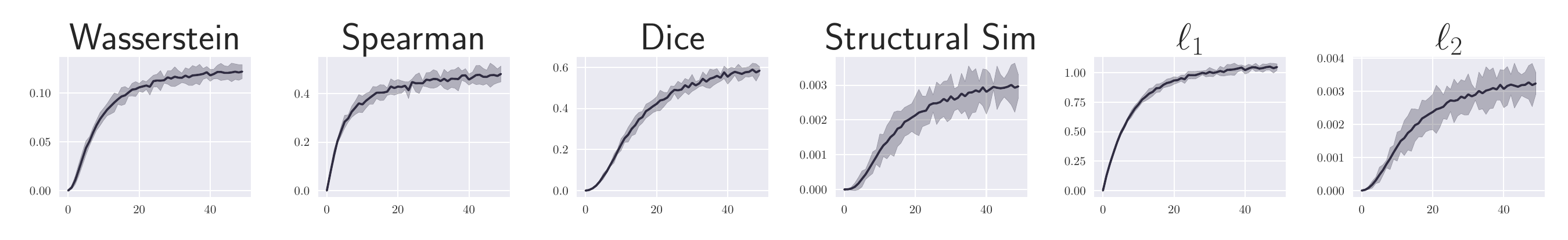}
  \caption{
    Distances with noisy images.
    The first line shows original noise-free image (left) and noisy copies computed by increasing $\sigma$. The second line shows the distances between each noisy image and the baseline image.
    }
  \label{dist:noise}
\end{figure}
Over the different distances tested, they all pass the sanity test : there is a monotonous positive correlation (as seen in Fig. \ref{dist:noise}). Although SSIM and $\ell_2$ have a higher variance.

One will nevertheless note the instability of the Dice score in cases where the areas of interest have a low surface area, as well as a significant computation cost for the Wasserstein distance. For all these reasons, we chose to stay in line with previous work using the absolute value of Spearman rank correlation.

\section{Additional results}
\label{ap:results}

\begin{table}[h]
    \centering
    \scalebox{0.85}{
        \begin{tabular}{llllllll}
        \toprule
        {Metrics} & IG & SG & SA & GI & GC & G+ & RI \\
        \midrule
        $\mu F$       & 0.11 & 0.31 & 0.23 & 0.10 & \textbf{0.91} & \underline{0.89} & 0.84 \\
        \mR           & 0.58 & 0.46 & 0.45 & 0.55 & \underline{0.72} & \textbf{0.82} & 0.56 \\
        \mC           & 0.11 & 0.15 & 0.15 & 0.09 & \textbf{0.64} & 0.49 & \underline{0.52} \\
        \bottomrule \\
        \end{tabular}
    }
    \caption{\fidelity, \consistency~and \representativity~score for ResNet-18 models on Cifar10. Higher is better. The first and second best results are respectively in \textbf{bold} and \underline{underlined}.}
    \label{tab:cifar_metrics}
\end{table}
\begin{table}[h]
    \centering
    \scalebox{0.85}{
        \begin{tabular}{llllllll}
        \toprule
        {Metrics} & IG & SG & SA & GI & GC & G+ & RI \\
        \midrule
        \mR           & 0.40 & \underline{0.42} & 0.41 & 0.41 & \textbf{0.67} & \textbf{0.67} & 0.39 \\
        \mC           & 0.31 & 0.18 & 0.18 & 0.23 & \underline{0.59} & \textbf{0.64} & 0.34 \\
        \bottomrule \\
        \end{tabular}
    }
    \caption{\consistency~and \representativity~score for ResNet-18 models on Eurosat. Higher is better. The first and second best results are respectively in \textbf{bold} and \underline{underlined}.}
    \label{tab:eurosat_metrics}
\end{table}
\begin{table}[h]
    \centering
    \scalebox{0.85}{
        \begin{tabular}{llllllll}
        \toprule
        {Metrics} & IG & SG & SA & GI & GC & G+ & RI \\
        \midrule
        \mR           & \textbf{0.90} & 0.36 & 0.30 & \textbf{0.90} & 0.77 & \underline{0.84} & 0.52 \\
        \mC           & \underline{0.37} & 0.13 & 0.10 & \underline{0.37} & \textbf{0.52} & 0.32 & \underline{0.37} \\
        \bottomrule \\
        \end{tabular}
    }
    \caption{\consistency~and \representativity~score for ResNet-18 models on Fashion-MNIST. Higher is better. The first and second best results are respectively in \textbf{bold} and \underline{underlined}.}
    \label{tab:fashion_metrics}
\end{table}

\begin{figure}[t!]
    \centering
    \includegraphics[width=0.4\textwidth]{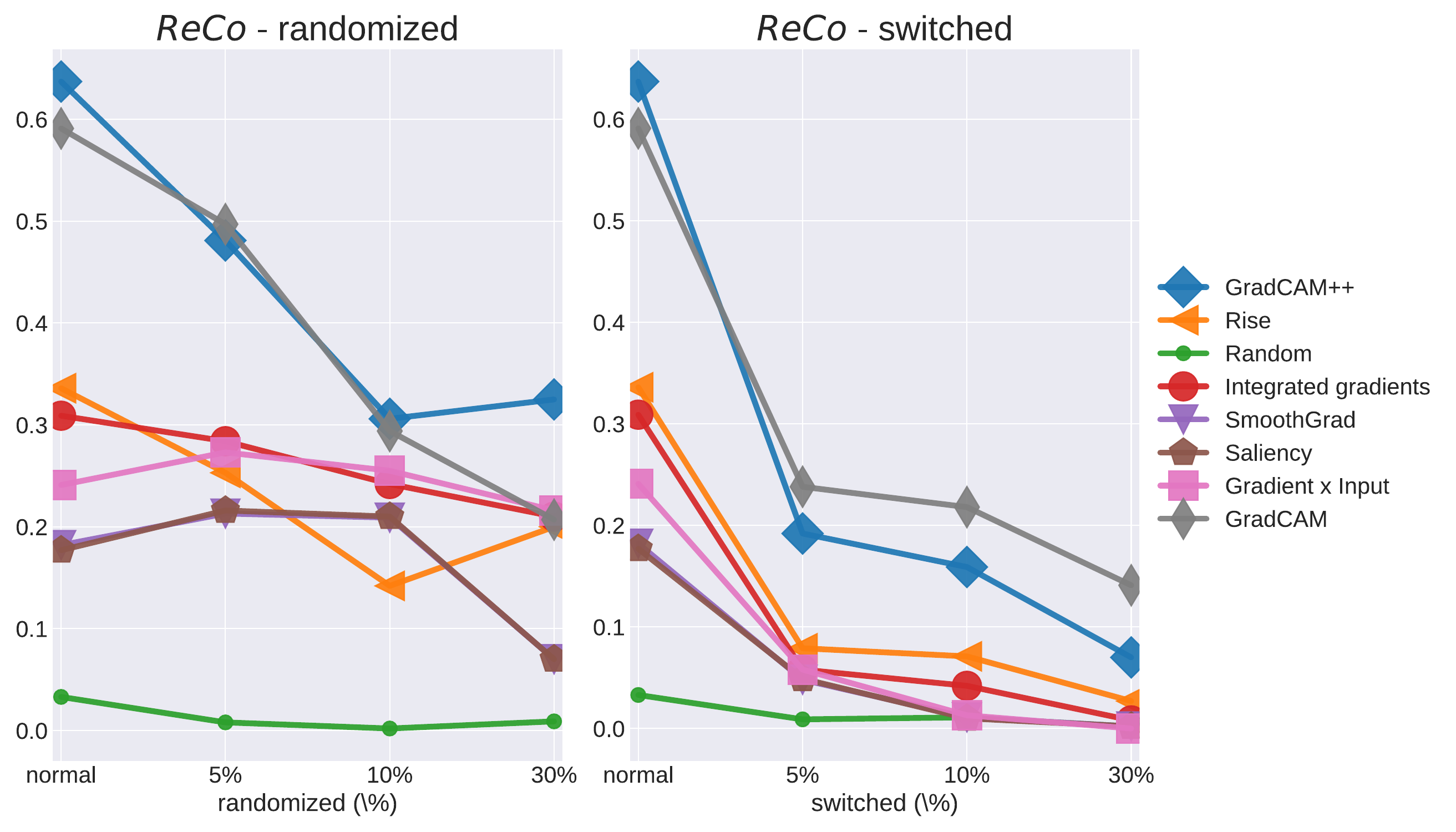}
    \includegraphics[width=0.4\textwidth]{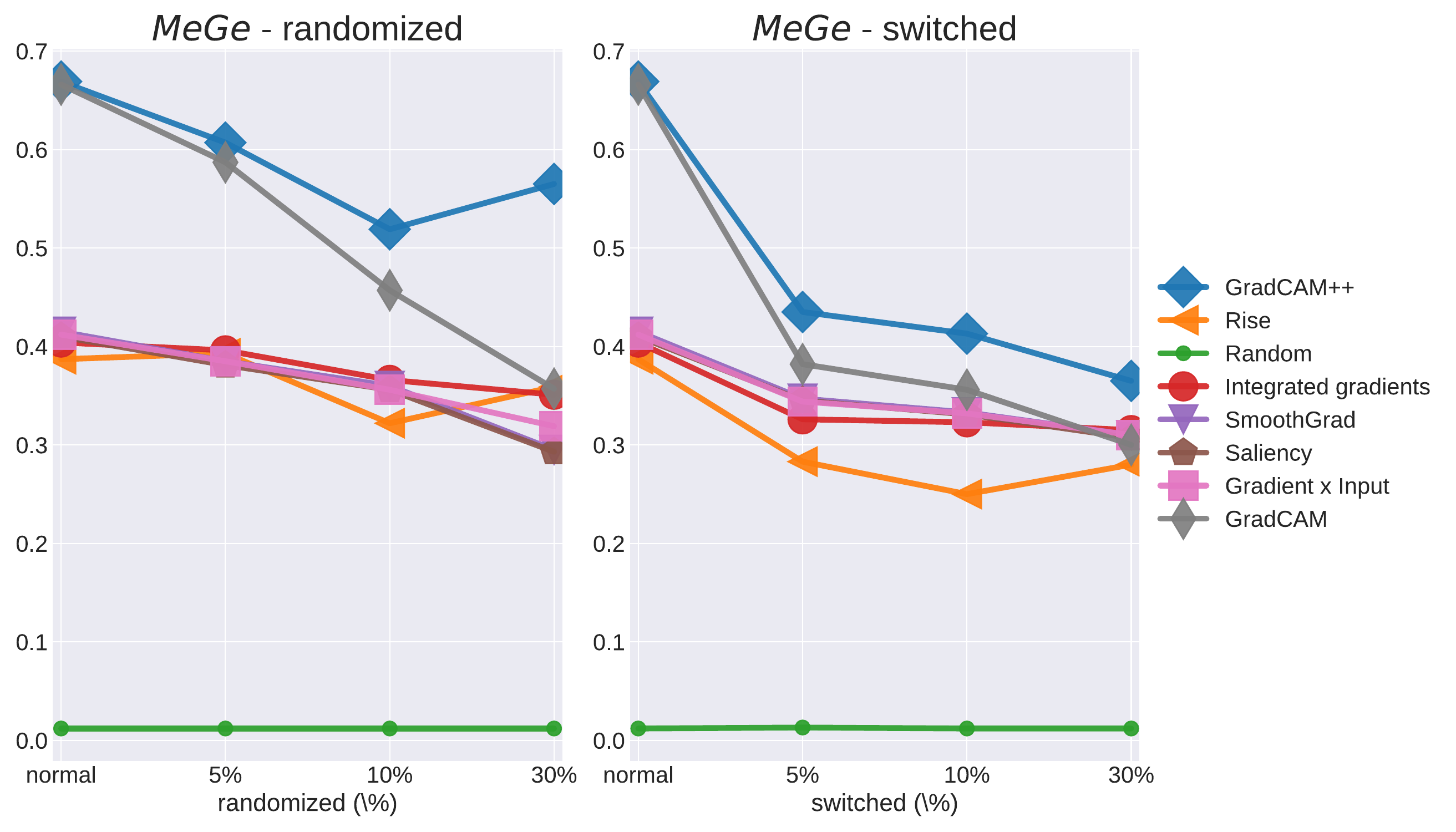}
    \caption{
    \textbf{Eurosat}~\mR~and \mC~scores for normally trained models (first point from the left), as well as for progressively randomized models and models trained with switched labels.}
\end{figure}
\begin{figure}[t!]
    \centering
    \includegraphics[width=0.4\textwidth]{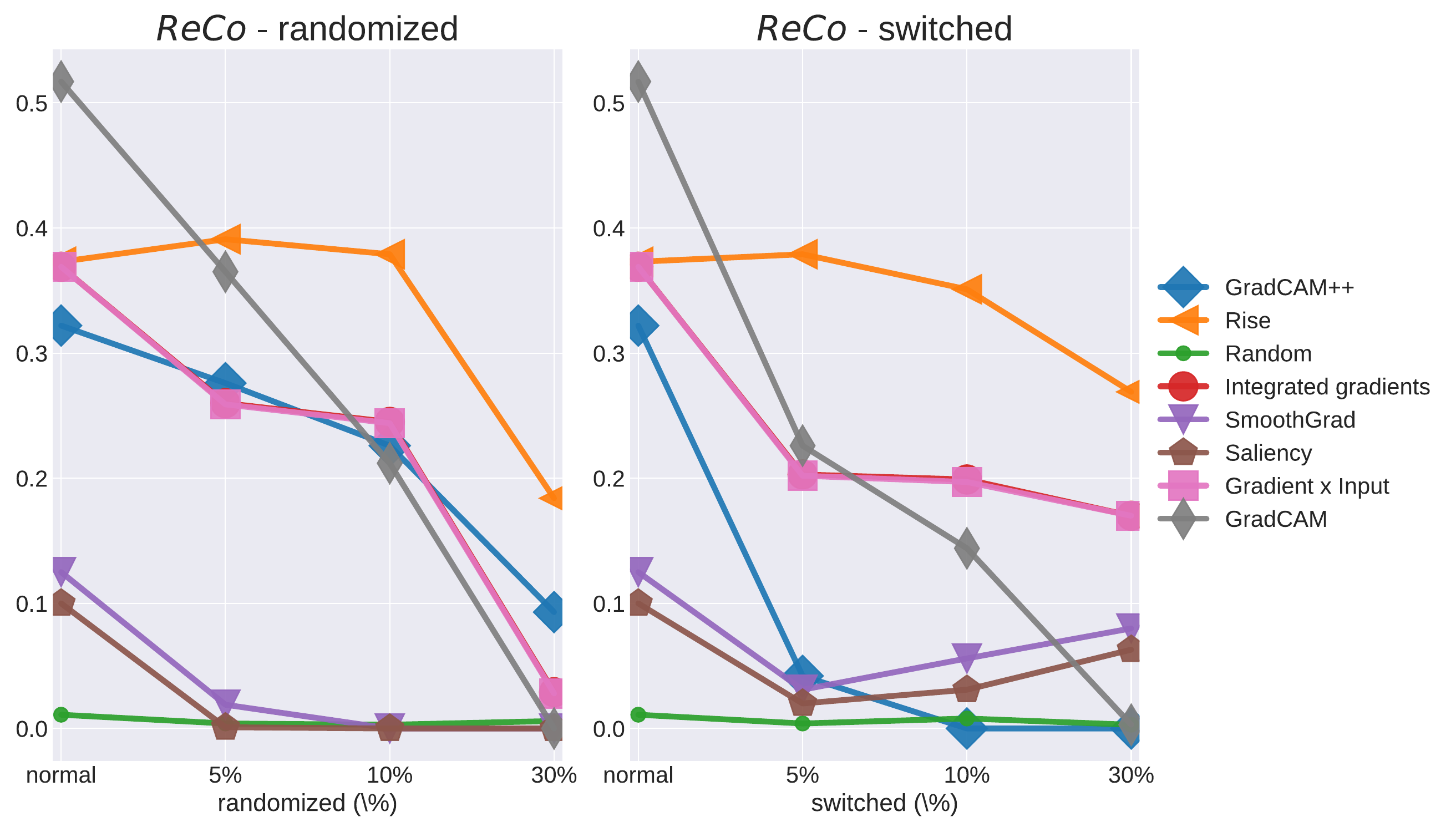}
    \includegraphics[width=0.4\textwidth]{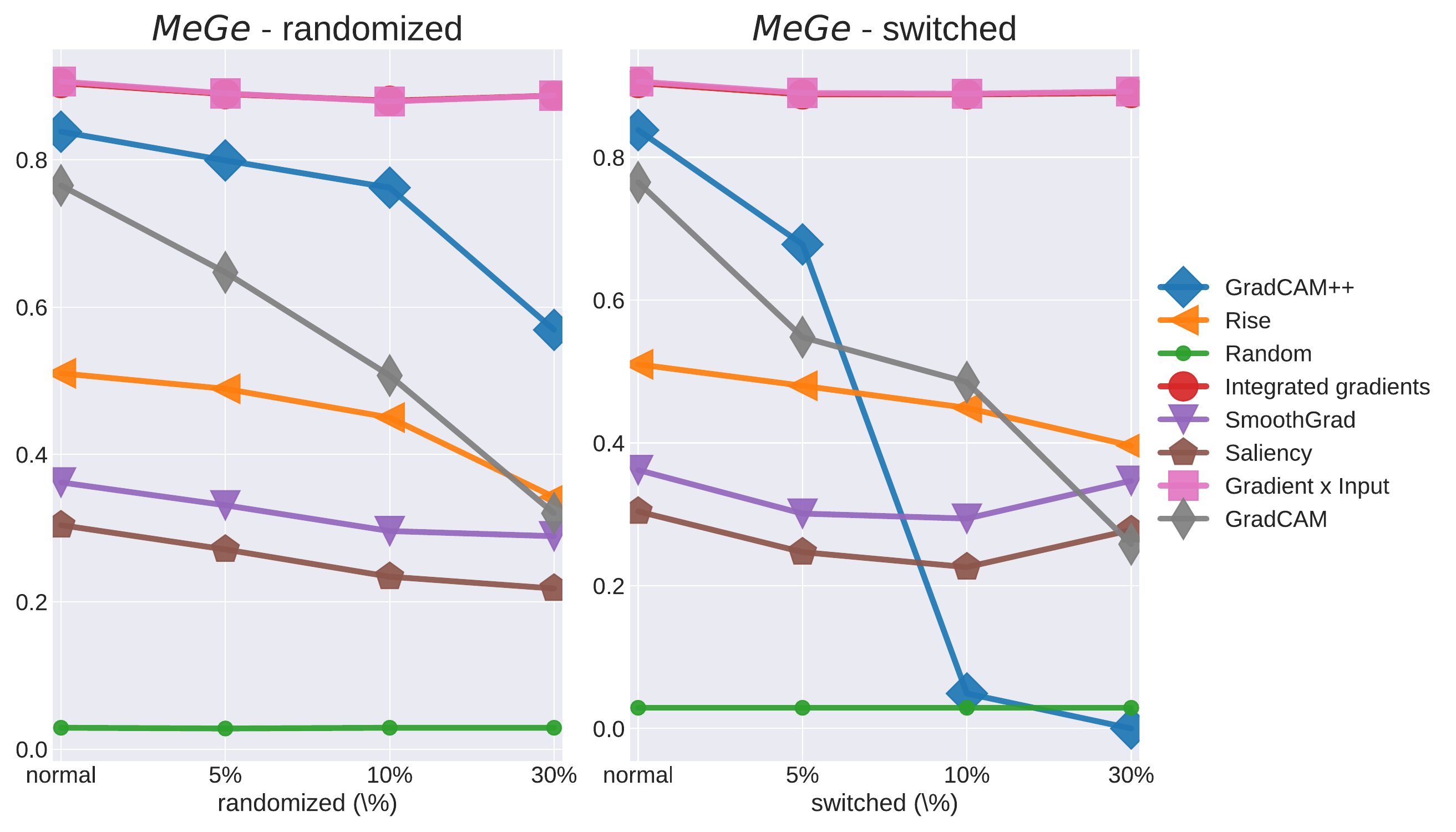}
    \caption{
    \textbf{Fashion-MNIST}~\mR~and \mC~scores for normally trained models (first point from the left), as well as for progressively randomized models and models trained with switched labels.}
\end{figure}

\begin{figure}[t!]
    \centering
    \includegraphics[width=0.5\textwidth]{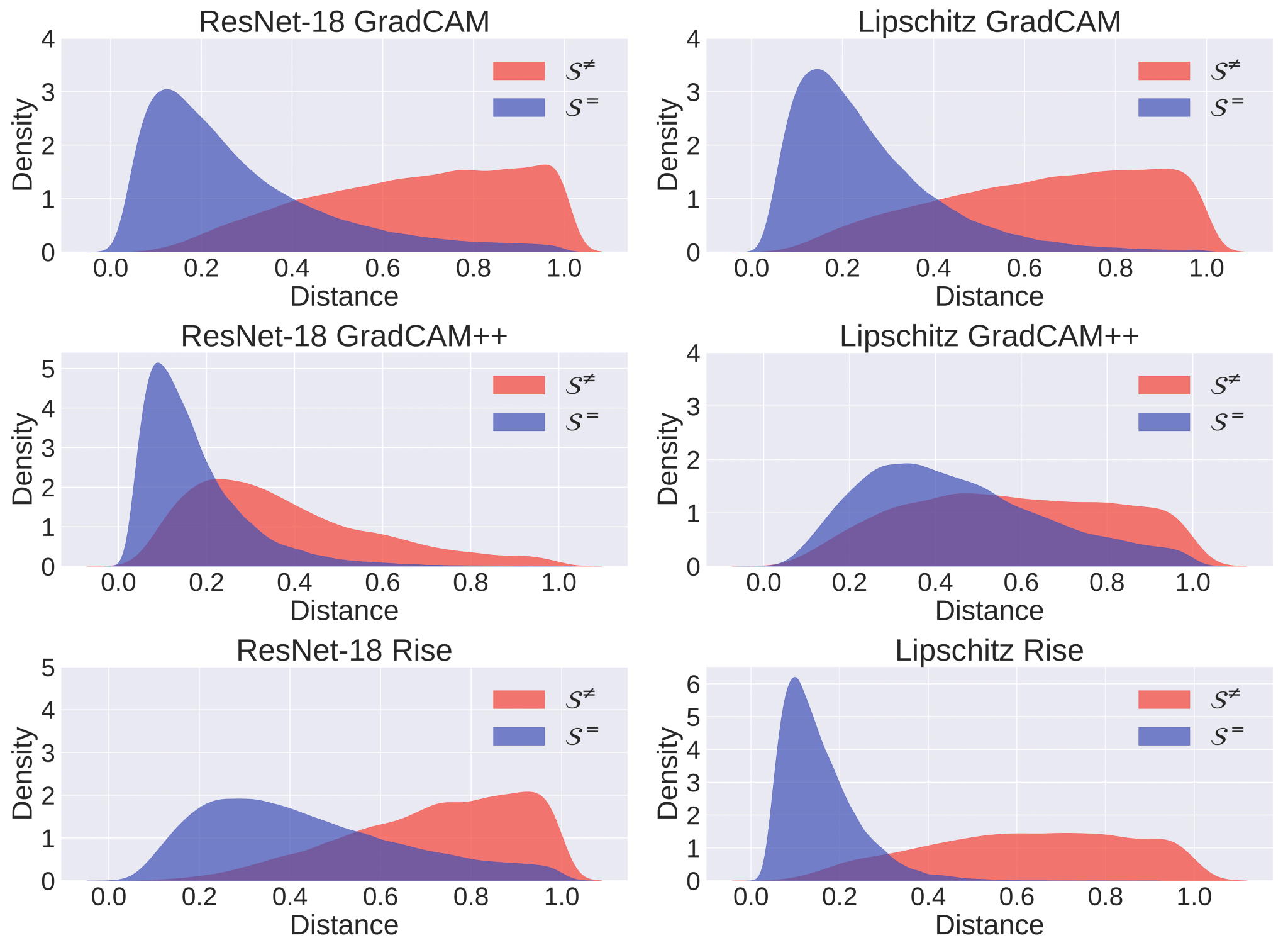}
    \caption{
    $\Sm$ and $\Sp$ for ResNet (left column) and 1-Lipschitz models (right column) on Cifar10.
    As explained in this paper, a clear separation between the $\Sm$ and $\Sp$ histograms is a sign of consistent explanations.
    }
    \label{fig:lipVsNormalDistrib2}
\end{figure}

\end{document}